\pdfoutput=1

\documentclass[11pt]{article}

\usepackage{acl2023}
\usepackage{times}
\usepackage{latexsym}

\usepackage[T1]{fontenc}

\usepackage[utf8]{inputenc}
\usepackage{natbib}  
\usepackage{caption} 
\usepackage{graphicx}
\usepackage{multirow,array}
\usepackage{bm}
\usepackage{bbm}
\usepackage{algorithm,algorithmicx}
\usepackage{amsmath}
\usepackage{amsthm}
\usepackage{verbatim}
\usepackage{booktabs}
\usepackage{amsfonts,amssymb} 
\usepackage[noend]{algpseudocode}

\newcommand{\secref}[1]{Section \ref{#1}}
\newcommand{\figref}[1]{Figure \ref{#1}}
\newcommand{\eqnref}[1]{Eq. (\ref{#1})}
\newcommand{\tabref}[1]{Table \ref{#1}}

\title{Low-Rank Prune-And-Factorize for Language Model Compression}
\author{Siyu Ren \hspace*{1cm} Kenny Q. Zhu\textsuperscript{\rm}\thanks{\hspace{2mm}The corresponding author.}\\
	Shanghai Jiao Tong University\\
	Shanghai, China\\
	roy0702@sjtu.edu.cn, kzhu@cs.sjtu.edu.cn}

\begin{document}
	\maketitle
	\begin{abstract}
		The components underpinning PLMs---large weight matrices---were shown to bear considerable redundancy. Matrix factorization, 
		a well-established technique from matrix theory, has been utilized to reduce the number of parameters in PLM. However, it fails to retain satisfactory performance under moderate to high compression rate. In this paper, we identify the \textit{full-rankness} of fine-tuned PLM as the fundamental bottleneck for the failure of matrix factorization and explore the use of network pruning to extract low-rank sparsity pattern desirable to matrix factorization. We find such low-rank sparsity pattern exclusively exists in models generated by first-order pruning, which motivates us to unite the two approaches and achieve more effective model compression.  We further propose two techniques: sparsity-aware SVD and mixed-rank fine-tuning, which improve the initialization and training of the compression procedure, respectively.  Experiments on GLUE and question-answering tasks 
show that the proposed method has superior compression-performance trade-off compared to existing approaches.
	\end{abstract}
	

	\section{Introduction}

Transformer-based~\cite{transformer} pre-trained language models~(PLMs)~\cite{bert,roberta} have shown superb performance on a variety of natural language processing tasks. These models are heavily over-parametrized~\cite{overpara} as they usually contain hundreds of millions of parameters, placing a severe burden on local storage, network transferring, runtime memory, and computation cost. Due to this disadvantage, the application of PLMs in low-resource scenarios is limited.

To alleviate this problem, recent studies~\cite{l0,svd} have attempted to compress 
PLMs by exploring and reducing the parameter redundancy in the weight matrices. Matrix factorization~(MF) , originated from linear algebra and matrix theory, is leveraged by modern deep learning towards achieving parameter efficiency. It works by decomposing large matrices into smaller sub-matrices with structural properties. The factorized sub-matrices serve as approximations of the original matrices while having fewer parameters.  \citet{svd} employ singular value decomposition~(SVD) for BERT compression with 2x compression rate and show 5\% drop in average GLUE~\cite{glue} performance compared to full BERT. The degradation is more evident under high compression rates~(\secref{sec:pilot_results}). Through a preliminary study, we identify the reason for the unsatisfactory performance of MF to be the \textit{full-rankness} of a fine-tuned language model. It inevitably causes information loss during the factorization process since the rank of sub-matrices has to be significantly smaller than the fine-tuned model to achieve parameter compression.

In an attempt to address this limitation of matrix factorization, we first explore the effect of network sparsification to produce subnetworks with the majority of weights set to zero. Ideally, we expect the subnetwork to contain low-rank sparse weight matrices and meanwhile preserve useful information for the end task. 
To this end, we conduct a systematic investigation into unstructured pruning~(UP) to study whether the resulting subnetworks exhibit the desirable low-rank property. From our experiments, we make the following important observations: (1) zero-order UP that only considers weight magnitude as pruning criterion produces subnetworks as full-rank as fine-tuned models; (2) first-order UP that incorporates gradient information into pruning decision is able to identify subnetworks that are both accurate and low-rank.

The above  findings motivate us to further explore the possibility of improving MF with UP.  Specifically, we design a sequential framework in which the first-order UP is executed prior to MF. In this way, the accurate low-rank 
subnetworks can be exploited by MF with minimal accuracy degradation while 
enjoying parameter and computation efficiency.

Moreover, we noticed that the vanilla SVD is not designed for 
sparse matrices because it penalizes the reconstruction error of 
each parameter equally~\cite{group}. Also, due to the reduced capacity, 
the joint re-training of low-rank sub-matrices may converge to 
solutions with lower generalization ability. To address the first problem, 
we propose sparsity-aware SVD, a weighted variant of SVD that 
better reconstructs unpruned~(hence more important) parameters. 
To address the second problem, we introduce mixed-rank fine-tuning, 
a regularized training scheme where the low-rank sub-matrices 
are randomly replaced with the sparse matrix from which they are factorized. Our contributions are  as follows:
\begin{itemize}
	\item Through a comprehensive preliminary study, 
	we discover a low-rank phenomenon in models obtained by first-order UP, 
	which highlights the possibility of a more efficient parametrization 
	of low-rank sparse matrices using low-rank factorization.
	\item Based on our findings, we design a sequential framework named  \textbf{L}ow-rank \textbf{P}rung-\textbf{A}nd-\textbf{F}actorize(LPAF) which makes high compression rate using matrix factorization possible.
	As further optimizations, we propose \textit{sparsity-aware SVD} which 
	prioritizes reconstruction of unpruned weights at initialization, 
	and \textit{mixed-rank fine-tuning} to compensate for the reduced capacity 
	during training.
	\item Comprehensive experiments on GLUE and 
	question-answering tasks show that our approach can achieve 
	a 2x-6x reduction in model size and FLOPs while retaining 99.8\%-96.2\% performance of the original BERT.
\end{itemize}

	\section{Background and Related Work}
\begin{figure}[t!]
	\centering
	\scalebox{0.154}{\includegraphics{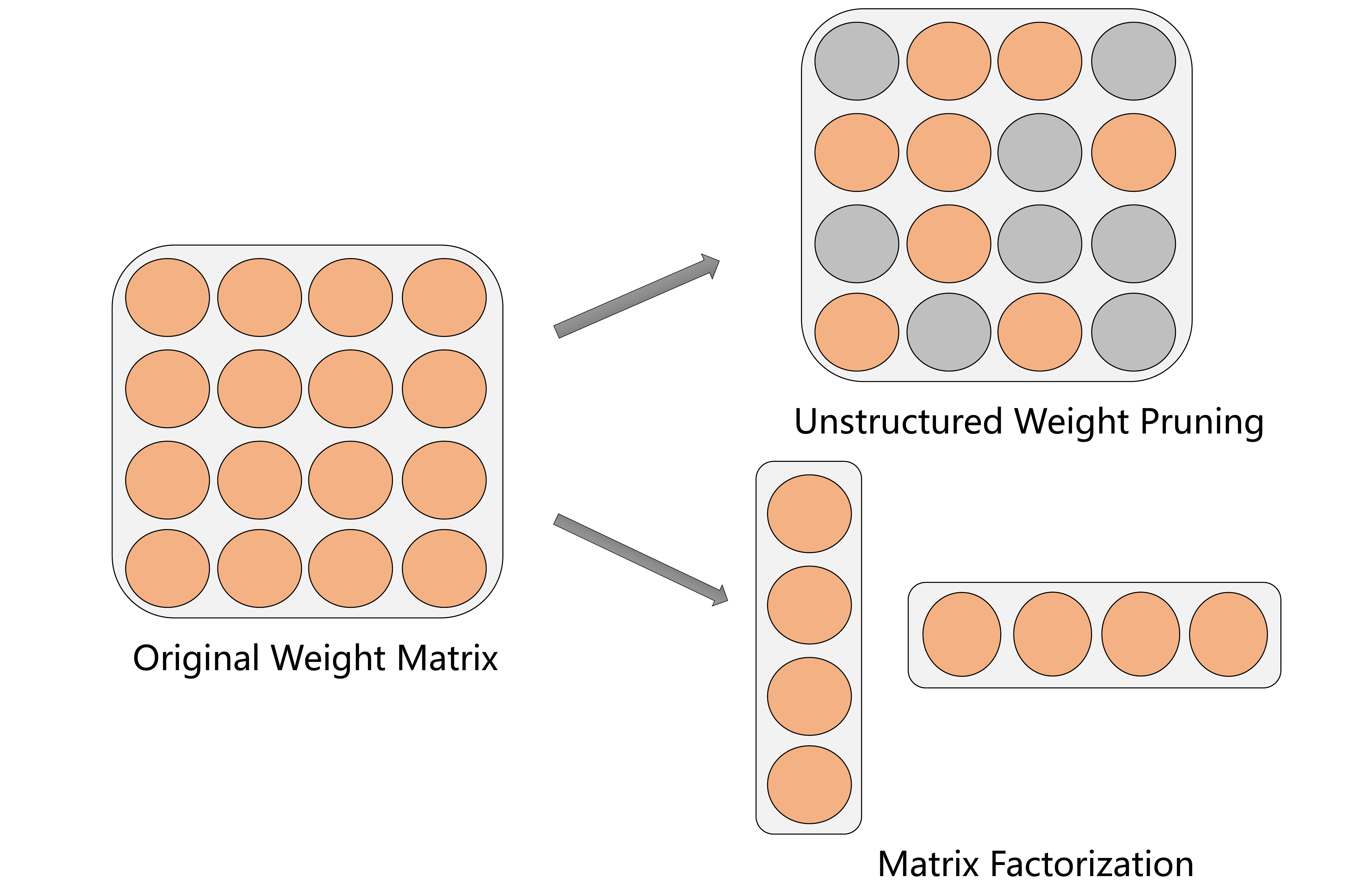}}
	\caption{Illustration of matrix factorization and unstructured pruning  on a single weight matrix.}
	\label{fig:intro}
\end{figure}

In this section, we present the necessary background knowledge about matrix factorization and unstructured pruning (\figref{fig:intro}).

\subsection{Matrix Factorization~(MF)}
\label{sec:lr}
Given the weight matrix $\bm{W}\in \mathbb{R}^{n\times m}$, matrix factorization~\cite{svd} decomposes it into sub-matrices with reduced total number of parameters to achieve model compression.  
It first uses singular value decomposition~(SVD) to obtain an equivalent 
form of $\bm{W}$ as the product of three matrices:
\begin{align}
	\bm{W}=\bm{U}\bm{\Sigma}\bm{V}^\mathrm{T}
\end{align}
where $\bm{U}\in \mathbb{R}^{n\times r}$, $\bm{\Sigma}\in  \mathbb{R}^{r\times r}$, $\bm{V}\in \mathbb{R}^{r\times m}$, and $r$ is the rank of matrix $\bm{W}$. $\bm{\Sigma}$ is a diagonal matrix of non-zero singular values $\{\sigma_1, \sigma_2,...,\sigma_r\}$ in descending order. Then, low-rank approximation with targeted rank $k$ is obtained by keeping the top-$k$ singular values in $\bm{\Sigma}$ as well as their corresponding column vectors in $\bm{U}$ and $\bm{V}$:
\begin{align}
	\bm{W}&\approx \bm{U}_{[:, :k]}\bm{\Sigma}_{[:k,:k]}\bm{V}_{[:, :k]}^{\mathrm{T}} =\bm{A}\bm{B}
	\label{eq:svd}
\end{align}
where $\bm{A}=\bm{U}_{[:,:k]}\bm{\Sigma}_{[:k,:k]}$ and $\bm{B}=\bm{V}_{[:,:k]}^{\mathrm{T}}$ are the two final sub-matrices of which the product is used to replace $\bm{W}$. After such factorization, the number of parameters is reduced from $nm$ to $k(n+m)$. Different compression rates can be achieved by varying the preserved rank $k$.

\subsection{Unstructured  Pruning~(UP)}
\label{sec:pruning}
Let $\bm{W}\in \mathbb{R}^{n\times m}$ denote a generic weight matrix in a PLM. In order to determine which elements in $\bm{W}$ are pruned, an importance score matrix $\bm{S}\in \mathbb{R}^{n\times m}$ is correspondingly introduced. The smaller $S_{i,j}$ is, the larger the probability of $W_{i,j}$ will be pruned. Given the importance scores, a pruning strategy $f_{prune}(\cdot)$ computes a binary mask matrix $\bm{M}\in \{0,1\}^{n\times m}=f_{prune}(\bm{S})$, 
and the forward process for an input $x$ becomes $y=(\bm{W}\odot\bm{M})x$, 
where $\odot$ denotes element-wise multiplication.

\paragraph{Zero-order Pruning~(UP$_{\text{zero}}$)} Zero-order pruning refers to the family of algorithms that only use the value of the weight as the importance measure.
For example, magnitude-based weights pruning~\cite{mag,chen2020lottery} adopts the absolute value of weight as importance score, i.e., 
$\bm{S}_{i, j}=|\bm{W}_{i, j}|$. The typical choice of $f_{prune}(\cdot)$ is to keep $v\%$  of weights with the largest importance scores:
\begin{align}
	\bm{M}_{i,j}=
	\begin{cases} 
		1, & \text{if }\bm{S}_{i,j}~\text{is in the largest }v\%\\
		0,  & \text{otherwise}  
	\end{cases}
	\label{eq:zero}
\end{align}

\paragraph{First-order Pruning~(UP$_\text{first}$)} Unlike zero-order pruning where $\bm{S}$ is directly derived from $\bm{W}$, first-order methods treat 
$\bm{S}$ as learnable parameters and jointly train it with model weights 
during fine-tuning. For example, SMvP~\cite{movement} and CAP~\cite{cap}
randomly initialize $\bm{S}$ and update it during the whole pruning process. The pruning strategy $f_{prune}(\cdot)$ is the same as in zero-order pruning~(\eqnref{eq:zero}).

Since the gradient of the thresholding function is 0 everywhere, straight-through estimator~\cite{st} is used as an approximation. The importance score $\bm{S}_{i,j}$ of $\bm{W}_{i,j}$ up to training step $T$ can be expressed as: 
\begin{align}
\bm{S}_{i,j}=-\sum_{t\le T}(\frac{\partial \mathcal{L}}{\partial \bm{W}_{i,j}})^{(t)} \bm{W}_{i,j}^{(t)}
\end{align}
where $\mathcal{L}$ is the loss function. The formulation is also equivalent to the first-order Taylor approximation of the change in $\mathcal{L}$ if $\bm{W}_{i,j}$ is zeroed out.

\paragraph{Sparsity Scheduler}
The proportion of remaining weights is controlled by the sparsity scheduler, here  we adopt the commonly used  cubic sparsity schedule to progressively reach target sparsity, i.e., $v_t$ at time step $t$ is derived by:
\begin{align}
	\begin{cases} 
		v_i & t\in [0, t_i) \\
		v_f+(v_i-v_f)(\frac{T-t_{f}-t}{T-t_f-t_i})^3 & t\in[t_i, T-t_f) \\
		v_f  & \text{otherwise}  
	\end{cases}
\end{align}
\label{eq:prune}
where $v_i=1.0$, $v_f$ is the final percent of remained parameters, $t_i$ and $t_f$ are the warmup and cool-down steps. $T$ is the total training steps. Moreover, we discard $\bm{M}$ and directly set $\bm{W}_{i,j}$ to zero if $\bm{S}_{i,j}^{(t)}$ is not in the top-$v_t$ at time step $t$. 
%



\section{Preliminary Study}
\label{sec:pilot}
In this section, we conduct a preliminary study on unstructured pruning  and matrix factorization
based on BERT-base and try to find answers to the following two questions: (1) How does matrix factorization perform under high compression rates? (2) Do subnetworks produced by unstructured pruning contain \textit{low-rank} sparsity patterns while preserving the majority of task accuracy?

\subsection{Experimental Setting}
\indent
\paragraph{Datasets}We use two tasks from GLUE benchmark~\cite{glue}, namely MRPC and RTE, as our evaluation testbeds. Both of them are formulated as classification problems.

\paragraph{Implementation Details} For matrix factorization, we follow the algorithm in \secref{sec:lr}. Specifically, we first fine-tune BERT-base on each downstream task following \citet{bert}. Then, we perform truncated SVD on weight matrices of each linear layer in the fine-tuned BERT and re-train the whole model to recover the lost accuracy. We select preserved rank $k$ from $\{390, 260, 130, 50\}$, which corresponds to $\{0.75, 0.50, 0.25, 0.10\}$ of BERT's parameters.

For unstructured  pruning, we evaluate both UP$_\text{zero}$ and UP$_\text{first}$. We set the value of $v_f$ from $\{0.75, 0.50, 0.25, 0.10\}$ to make a direct comparison to matrix factorization.

\subsection{Results and Analysis}
\label{sec:pilot_results}


\paragraph{Accuracy Preservation} 
The variation of task accuracy with respect to the remaining parameters is illustrated 
in the top half of \figref{fig:pre}. Under a small compression rate, i.e., 
$75\%$  parameters remaining, all examined methods can retain $\ge 97\%$ performance 
of BERT-base across all tasks. Under moderate compression rate, i.e., $50\%$ parameters remaining, UP$_\text{zero}$ and SVD start to show obvious declines. 
When more extreme compression rates are pursued, e.g., $25\%$-$10\%$ parameters 
remaining, SVD exhibits the most drastic performance drops compared to UP methods. 
On the contrary,  UP$_\text{first}$ still retains $\sim 97.6\%$ of BERT's performance.
UP$_\text{zero}$ lags behind UP$_\text{first}$ by a large margin under high sparsity. This indicates that magnitude alone cannot be used to quantify a weight's 
contribution because even a small weight can yield a huge influence on the model 
output due to the complicated compositional nature of neural networks. 
In contrast, the importance criterion of UP$_\text{first}$ directly reflects the 
sensitivity of the model's training loss w.r.t. each weight and is therefore more 
accurate.

\begin{figure}[t]
	\centering
	\scalebox{0.175}{\includegraphics{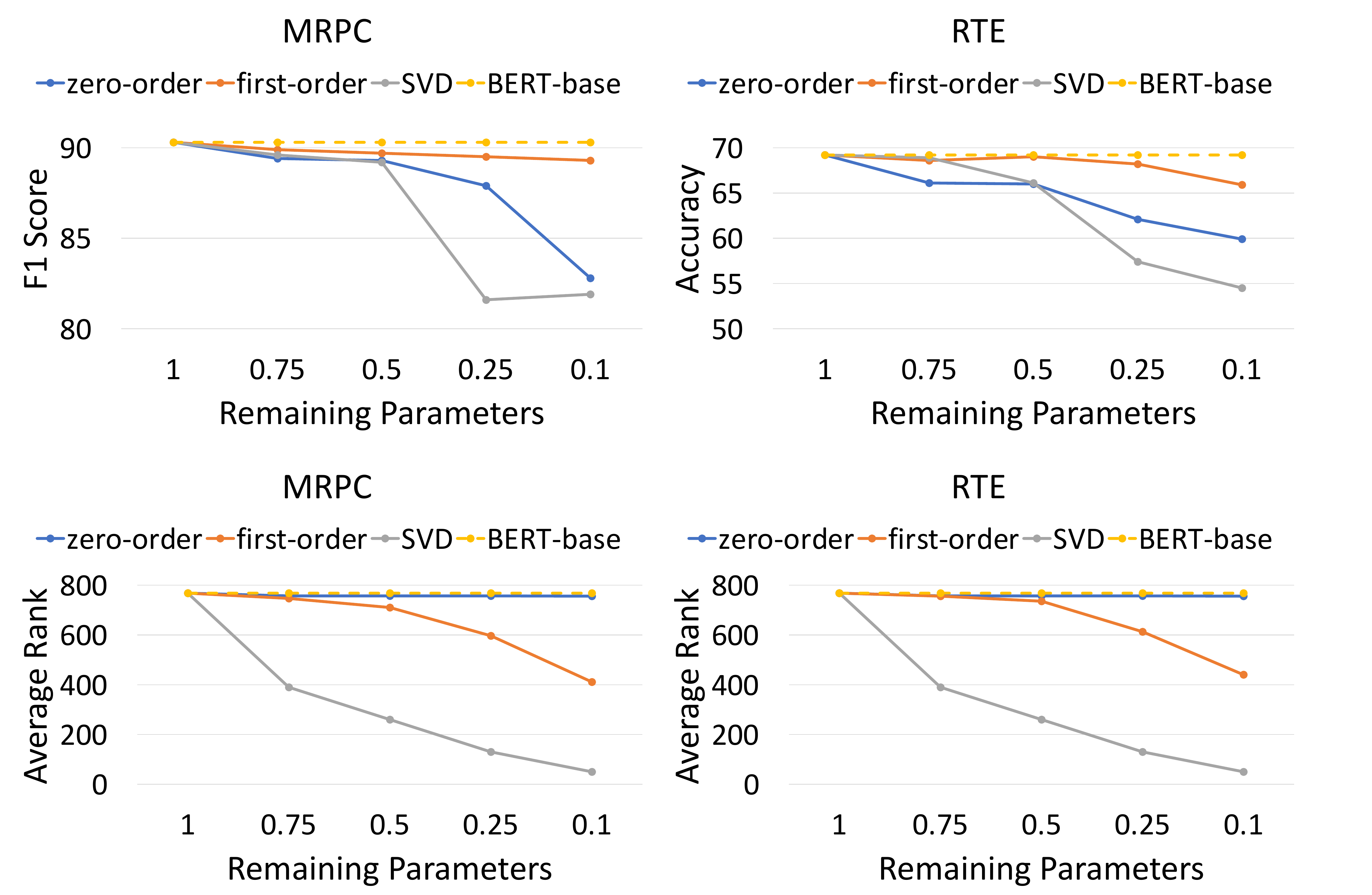}}
	\caption{Task accuracy~(top half) and average matrix rank~(bottom half) 
v.s. percentage of original parameters retained. 
The dashed line indicates the performance/rank upper bound by fine-tuning the full-scale BERT-base model. Results on more datasets are deferred to Appendix \ref{sec:A}.}
	\label{fig:pre}
\end{figure}

\begin{figure}[t]
	\centering
		\scalebox{0.50}{\includegraphics{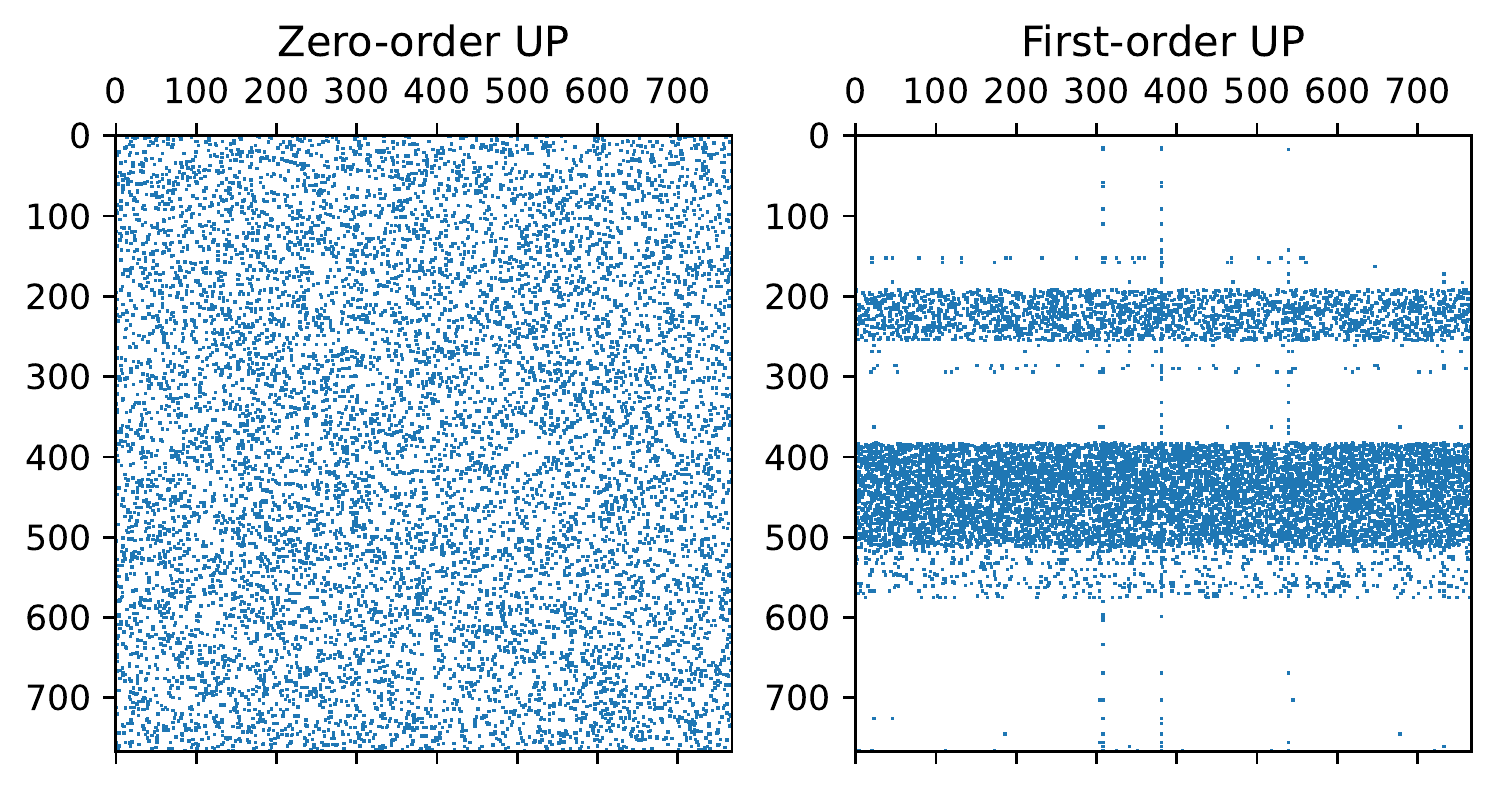}}
	\caption{Sparsity patterns of the same 768x768 weight matrix  pruned by UP$_\text{zero}$~(left) and UP$_\text{first}$~(right) on MRPC with $10\%$ of
the parameters remaining.}
	\label{fig:pattern}
\end{figure}

\paragraph{Rank} 
Considering the inferior accuracy of SVD, we hypothesize that the weight matrices of fine-tuned BERT are high-rank, 
hence leading to a large approximation error when $k$ is small. The bottom half of \figref{fig:pre} inspects the average rank of weight matrices. We can see that the weight matrices in fine-tuned BERT-base are nearly full-rank, which explains the inefficacy of SVD when $k$ is small. We also plot the rank-parameter curve of UP methods. For UP$_\text{zero}$, it produces sparse matrices that are 
as high-rank as densely fine-tuned BERT even when $90\%$ weights are set to zero. In contrast, UP$_\text{first}$  produces sparse patterns whose rank monotonically decreases as more weights are pruned. To gain more insights into this phenomenon, we visualize the weight matrix pruned by UP$_\text{zero}$ and UP$_\text{first}$ in \figref{fig:pattern}. Though both are designed without structural bias,  unlike UP$_\text{zero}$, UP$_\text{first}$ learns to remove entire rows from the weight matrix and 
the resulting matrix enjoys a low-rank characteristic.

\begin{figure}[th]
	\centering
	\scalebox{0.142}{\includegraphics{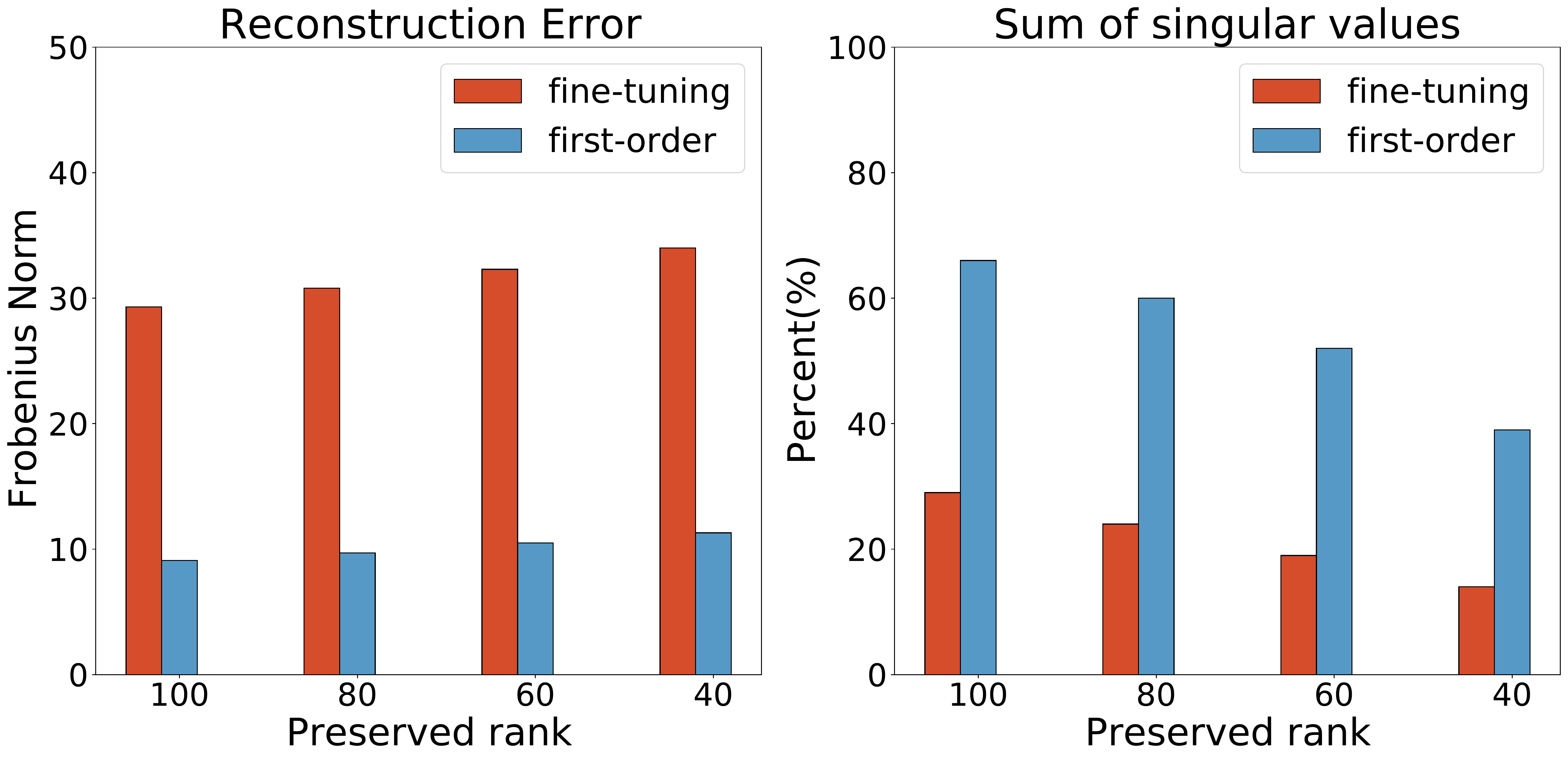}}
	\caption{Quantitatively measuring approximation quality via reconstruction error~(left) and cumulative sum of singular values~(right) on MRPC.}
	\label{fig:norm}
\end{figure}

\paragraph{The Idea}
The key insight is: factorizing a high-rank matrix into low rank sub-matrices
loses significant quantity of useful information, but factorizing a low-rank matrix into low rank sub-matrices
doesn't lose as much information. Our design is based on this insight. 
As a sanity check of its feasibility, we quantitatively measure the 
quality of low-rank approximation with various preserved ranks $k$. 
\figref{fig:norm} shows that given a specific $k$, 
the sum of top-$k$ singular values of matrices produced by UP$_\text{first}$ takes a much larger portion of total values than fine-tuning, suggesting that we can reserve more information of low-rank sparse matrix given the same $k$. The reconstruction error~(measured by Frobenius norm) of UP$_\text{first}$ is also significantly lower, implying a higher approximation quality. We thus expect that low-rank matrix factorization on low-rank sparse models to effectively combine: 
(1) the good performance of first-order UP; 
(2) direct memory and computation reduction by MF.

\section{LPAF: Low-rank Prune-And-Factorize}
\label{sec:approach}
Here we formally propose the LPAF~(\textbf{L}ow-rank \textbf{P}rune-\textbf{A}nd-\textbf{F}actorize) framework 
for language model compression. In addition, 
we propose two optimizations in the 
initialization and training of the compression process.

\subsection{The Overall Workflow}
\label{sec:ptf}
Given a pre-trained language model $T$ and a downstream task with training set $D=\{(x_i, y_i), i=1,2,...M\}$, LPAF consists of three steps to realize model compression: 
\begin{itemize}
	\item Step-1: obtaining the low-rank sparse model $T_{\text{sparse}}=\text{UP}_\text{first}(T,D, v)$. $v$ is the percent of remained parameters after pruning.
	\item Step-2:  performing matrix factorization on each weight matrix~(excluding the embedding layer) in $T_{\text{sparse}}$ and  obtain its low-rank factorized form $T_{\text{factorized}}$. 
	\item Step-3:  re-training $T_{\text{factorized}}$ on $D$ using task-specific loss function until convergence. 
\end{itemize}

Next, we present two novel optimizations, namely \textit{sparsity-aware SVD} and \textit{mixed-rank fine-tuning}, that improves the matrix factorization and fine-tuning process in step 2 and step 3 respectively.

\subsection{Optimization 1: Sparsity-aware SVD}
\label{sec:sasvd}
SVD has been shown~\cite{bestsvd} to provide the optimal rank-$k$ approximation to $\bm{W}$ with respect to the Frobenius norm:
\begin{align}
	\nonumber
	\min_{\bm{A},\bm{B}} ||\bm{W}-&\bm{A}\bm{B}||_{F}=\min_{\bm{A},\bm{B}} \sum_{i,j}(\bm{W}_{i,j}-(\bm{AB})_{i,j})^2 \\
	& \text{s.t.}~~~~\text{rank}(\bm{AB})=k
\end{align}
It is a generic factorization method in that it is applicable to any matrix $\bm{W}$ by penalizing the reconstruction error of each individual weight equally. 

In our case, $\bm{W}$ is a sparse matrix from $T_{\text{sparse}}$ in which the majority of weights are set to zero by the pruning algorithm $P$. These zero weights are deemed to have less impact on the task performance compared to the retained~(unpruned) weights. However, the vanilla SVD treats each weight equally without considering the inherent sparseness of $W$, thus may be sub-optimal for preserving useful information in $W$ about the end task.
To address this issue, we propose sparsity-aware SVD which considers different priorities of parameters and weighs the individual reconstruction error based on its importance score $\bm{S}_{i,j}$:
\begin{align}
	\min_{\bm{A},\bm{B}} \sum_{i,j}&\bm{S}_{i,j}(\bm{W}_{i,j}-(\bm{AB})_{i,j})^2~~~\\\
	 & \text{s.t.}~~\text{rank}(\bm{AB})=k
	\label{eq:sasvd}
\end{align}
In this way, parameters that are more important can be better reconstructed, hence retaining more task performance from $T_{\text{sparse}}$ at initialization. Nevertheless, \eqnref{eq:sasvd} does not have a closed form solution~\cite{weightedsvd,hsu2021language} when each $\bm{W}_{i,j}$ has its own weight. We therefore resort to a simplification by letting the same row of $\bm{W}$ share the same importance. The importance for row $i$ is given by $\hat{\bm{S}}_{i}=\frac{\sum_{j}\bm{S}_{i,j}}{\sum_{n}\hat{\bm{S}}_{n}}$. Let $\hat{\bm{I}}=diag(\hat{\bm{S}}_1,\hat{\bm{S}}_2,...,\hat{\bm{S}}_{n})$ denote a diagonal matrix,  \eqnref{eq:sasvd} is now converted to:
\begin{align}
	&\min_{\bm{A},\bm{B}}||\hat{\bm{I}}\bm{W}-\hat{\bm{I}}\bm{A}\bm{B}||_F~~~~
	\\
	& \text{s.t.}~~\text{rank}(\bm{AB})=k
\end{align}
This essentially amounts to applying rank-$k$ SVD upon $\hat{\bm{I}}\bm{W}$, i.e., $\hat{\bm{I}}\bm{W}=\hat{\bm{U}}\hat{\bm{\Sigma}}\hat{\bm{V}}^\mathrm{T}$. Then the solution of $\bm{A}$ and $\bm{B}$ can be analytically obtained by:
\begin{align}
	\bm{A} &= \hat{\bm{I}}^{-1}\hat{\bm{U}}_{[:,:k]}\hat{\bm{\Sigma}}_{[:k,:k]},\bm{B}=\hat{\bm{V}}_{[:,:k]}^{\mathrm{T}}
\end{align}

\subsection{Optimization 2: Mixed-rank Fine-tuning}
Recall that the last step of LPAF is to fine-tune $T_{\text{factorized}}$ on the training set $D$. This process has been proven essential to regain the performance lost during factorization~\cite{svd}. However, during the experiments, we observe the performance of fine-tuned $T_{\text{factorized}}$ still slightly lags behind $T_{\text{sparse}}$ given a similar parameter budget. 
We posit that, due to the reduced capacity~(less trainable parameters) and model-level approximation error incurred by low-rank factorization, joint fine-tuning of low-rank matrices may converge to sub-optimal solutions with lower generalization ability. To mitigate this problem, we propose mixed-rank fine-tuning, a regularized scheme for training low-rank matrices.

Let $\{(\bm{A}\bm{B})_i, i=1,2...,N\}$ denotes all low-rank matrices in $T_{\text{factorized}}$. During training, for each $(\bm{A}\bm{B})_i$, we sample a binary Bernoulli random variable $z_i\sim \text{Bernoulli}(p)$, where $p$ is a global hyper-parameter. Then, the local computation process involving $(\bm{A}\bm{B})_i$ is modified to:
\begin{align}
	\bm{x}_{out} = (1-z_i)*(\bm{A}\bm{B})_i \bm{x}_{in} + z_i * \bm{W}_i\bm{x}_{in} 
\end{align}
where $\bm{W}_i$ is the sparse matrix in $T_{\text{sparse}}$ from which $\bm{A}_i$ and $\bm{B}_i$ are derived. In this way, the low-rank matrices can further 
benefit from gradient-level regularization from  $T_{\text{sparse}}$, 
thus reducing the generalization gap. The hyper-parameter $p$ is controlled by a scheduler. We implement it such that $p$ is linearly decayed from an initial value $p_{\text{init}}$ to zero by a constant step size $d$:
\begin{align}
	p = \text{max}(0, p_{\text{init}}-d*t)
\end{align}
As $p$ decreases, $\bm{W}_i$ is gradually substituted by low-rank sub-matrices $(\bm{AB})_i$. When $p$ reaches zero, the training enters the phase of standard fine-tuning. To further mitigate the training instability brought by sampling, we let each input go through the forward pass twice with different $\bm{z}^1=\{z_i^1\}_{i=1}^{N}$ and $\bm{z}^2=\{z_i^2\}_{i=1}^{N}$, and impose a consistency objective on the two outputs to promote stability:

\begin{align}
	\mathcal{L}_{c}=\mathcal{D}(y_{\bm{z}^1}, y_{\bm{z}^2})
\end{align}
where $\mathcal{D}$ can be the KL divergence for classification tasks and the MSE loss for regression tasks.


	\section{Experiments}
In this section, we present the experiments of LPAF for language model compression. We compare with state-of-the-art compression methods and perform detailed analysis of the results to provide guidance under different resource budgets.

\subsection{Experimental Setup}
\subsubsection{Datasets} 
We evaluate our approach on tasks from GLUE benchmark~\cite{glue}, SQuAD v1.1~\cite{qnliandsquad}, and SQuAD v2.0~\cite{squadv2.0} question-answering tasks. GLUE tasks include RTE, CoLA, SST-2~\cite{sst2}, MRPC, QQP, QNLI~\cite{mrpc}, and MNLI~\cite{mnli}.


\subsubsection{Baselines} We compare LPAF as well as its three ablated versions that remove each of the three steps in \secref{sec:ptf} against four categories of methods with a perceivable reduction in model size and computation.

\textbf{Pre-training Distillation. }DistilBERT~\cite{distilbert}, and TinyBERT~\cite{tinybert} are two pre-training distillation models using unlabeled corpus followed by task-specific fine-tuning.

\textbf{Task-specific Distillation. }PKD~\cite{pkd} extends KD by intermediate feature matching;
Theseus~\cite{theseus} proposes a progressive module replacing method for knowledge
distillation; CKD~\cite{CKD} transfers the contextual knowledge via word relation and layer transforming relation; MetaDistil~\cite{metadistil} uses meta-learning for training the teacher to better transfer knowledge to the student.

\textbf{Structured Pruning.} 
Iterative structured pruning~(ISP)~\cite{isp} removes attention heads and neurons in FFN layer with the lowest sensitivity in an iterative manner; FLOP~\cite{flop} adaptively removes rank-1 components of weight matrices during training; Block Pruning~(BP$_{\text{hybrid}}$)~\cite{block} shares pruning decisions for each 32x32 weight blocks in attention layer and for each row/columns in FFN layer; CoFi~\cite{sp-l0} jointly prunes attention heads, neurons, hidden dimension, and entire MHA/FFN layer via Lagrangian multipliers.

\textbf{Matrix Factorization. }SVD$_{\text{Ft}}$~\cite{svd} applies truncated SVD on a densely fine-tuned BERT and re-trains the factorized model to recover accuracy loss.

\begin{table}[t]
	\scriptsize
	\centering
	\begin{tabular}{l|cc|cc}
		\toprule
		& \multicolumn{2}{c|}{\textbf{\% of Parameters}} & \multicolumn{2}{c}{\textbf{FLOPs}} \\ 
		\midrule
		Task         & \multicolumn{2}{c|}{All}          & GLUE    & SQuAD v1.1/v2.0    \\ 
		\midrule
		BERT-base    & \multicolumn{2}{c|}{100\%}          & 7.4G      & 35.4G         \\
		\midrule
		LPAF-260     & \multicolumn{2}{c|}{50\%}           & 3.7G      & 16.1G         \\
		LPAF-130     & \multicolumn{2}{c|}{25\%}           & 1.9G      & 10.3G         \\
		LPAF-80      & \multicolumn{2}{c|}{16\%}           & 1.3G      & 7.9G          \\
		\bottomrule
	\end{tabular}
	\caption{Percentage of parameters and FLOPs. 
	}
	\label{table:stats}
\end{table}

\begin{table*}[t]
	\centering
	\scriptsize
	\begin{tabular}{l|ccccccc}
				\toprule
		Task &\textbf{RTE~(2.5K)} &\textbf{MRPC~(3.7K)} &\textbf{CoLA~(8.5K)} & \textbf{SST-2~(67K)}                                                  & \textbf{QQP~(364K)}                                                    & \textbf{QNLI~(105K)}                                                & \textbf{MNLI~(393K) }                                                    \\
				\midrule
		\% Params.   &50\% ~25\%  ~16\% &50\% ~25\%  ~16\% &50\% ~25\%  ~16\%            &50\% ~25\%  ~16\%             &50\% ~25\%  ~16\%            &50\% ~25\%  ~16\%            &50\% ~25\%  ~16\%        \\
		\hline
		DistilBERT  &65.0~ 61.0~ 56.3 &85.8~ 77.0~ 72.5 &51.3~ 32.1~ 21.1    & 90.0 ~88.9 ~86.4          & 90.8~ 89.4~ 88.0          & 86.0~ 83.8~ 81.6         & 81.7 ~76.4 ~71.3         \\
		TinyBERT    &67.7~ 67.2~ 64.6 &86.3~ 85.3~ 78.2 &53.8~ 33.3 ~21.3   & 92.3 ~89.8 ~88.0          & 90.5~ 90.0~ 88.7          & 89.9 ~87.7~ 84.5          & 83.1~ 80.6~ 77.4           \\
		\hline
		PKD  &65.5~ 59.2~ 53.8 &81.9~ 76.2~ 71.3 &45.5~ 22.0~ 19.1   & 91.3~ 88.1 ~87.2          & 88.4~ 88.5 ~87.5          & 88.4~ 82.7~ 78.0          & 81.3 ~75.7~ 72.7                                                  \\
		Theseus &65.6~ 62.1~ 58.8 &86.2~ 77.2~ 72.8 &51.1~ 17.9~ 17.6  & 91.5~ 88.5~ 86.1          & 89.6 ~89.0 ~86.0          & 89.5 ~85.0~ 80.3          & 82.3~ 76.4 ~73.5                                         \\
		CKD   &67.3~ 66.5~ 60.8 &86.0~ 81.1~ 76.6 &55.1~ 40.1~ 32.9   & \textbf{93.0} ~89.8~ 88.7          & 91.2~ 90.1~ 88.9          & 90.5~ 87.0 ~84.9          & 83.6 ~79.0~ 76.8                                \\
		MetaDistil  &69.0~ 66.7~ 61.0 &86.8~ 81.8~ 77.3 &\textbf{56.3}~ 33.6~ 24.3    & 92.3 ~88.9~ 87.0          & 91.0~ 88.9~ 86.9          & 90.4~ 86.8 ~84.9          & 83.5 ~79.5~ 76.8                             \\
		\hline
		ISP  &66.4~ 65.0~ 63.9 &86.1~ 83.6~ 82.8 &55.3~ 45.6~ 31.0   & 90.6~ 90.4 ~89.4          & 90.8~ 90.1 ~89.3          & 90.5~ 88.7~ 87.2          & 83.2 ~81.9~ 80.8                                                  \\
		FLOP &66.1~ 58.5~ 56.0 &82.1~ 80.1~ 78.4 &49.1~ 35.3~ 28.6   & 91.4~ 89.7~ 89.4          & 91.1~ 90.1~ 89.1          & 90.5~ 88.5~ 87.1          & 82.6 ~79.9~ 79.4                                                  \\
		BP$_{\text{hybrid}}$ &66.4~ 64.3~ 63.9&84.1~ 83.8~ 81.1 &50.0~ 37.3~ 35.4   & 90.8~ 89.8~ 89.2          & 90.8~ 90.1~ 89.8          & 90.2~ 88.7~ 88.1          & 83.2 ~80.6~ 80.1                                                  \\
		CoFi  &\textbf{69.3}~ 66.4~ 66.4 &84.6~ 84.3~ 83.6 &51.8~ 44.1~ 30.3   & 91.6~ 89.7 ~89.2          & 91.0~ 90.2 ~89.9          & 90.8~ 88.8~ 87.6          & 83.5 ~80.8~ 80.5                                                  \\
		\hline
		SVD$_{\text{Ft}}$  &62.1~ 60.3~ 55.6 &79.9~ 70.1~ 70.0&44.9~ 26.6~ 18.0    & 90.8 ~88.9~ 85.3         & 91.3 ~90.0~ 87.9          & 91.0 ~86.1~ 83.8          & 83.0~ 79.9~ 76.6         \\
		LPAF~(ours)  &68.2~ \textbf{68.0}~ \textbf{67.9} &\textbf{86.8}~ \textbf{86.5}~ \textbf{86.0} &55.5~ \textbf{48.5~ 42.8}     & 92.4~ \textbf{90.7~ 89.7} & \textbf{91.5}~ \textbf{90.4}~ \textbf{90.1} & \textbf{91.3~ 89.3~ 88.6} & \textbf{84.6}~ \textbf{82.6}~ \textbf{81.7}  \\
		~~-w/o Step-1   &64.2~ 32.1~ 21.1 &82.1~ 32.1~ 21.1 &49.0~ 32.9~ 18.2    &91.2~ 89.9 ~88.4 &91.3~ 90.3~ 89.7 &91.2~ 87.8~ 84.8 &83.3~ 82.0~ 79.6  \\
		~~-w/o Step-2  &65.3~ 32.1~ 21.1 &86.0~ 32.1~ 21.1   &52.0~ 48.0~ 41.0   &91.2~ 89.2~ 88.8 &91.2~ 90.2~ 90.0 &90.9~ 89.0~ 87.9 &83.4~ 82.4~ 81.5 \\
		~~-w/o Step-3   &65.0~ 32.1~ 21.1 &84.8~ 32.1~ 21.1 &52.9~ 48.2~ 42.2     &91.4~ 89.5~ 88.8 &91.1~ 90.3~ 89.9 &91.1~ 88.9~ 88.1 &83.0~ 81.3~ 81.0 \\
\midrule
		BERT-base  &~~69.2~~~ &~~86.4~~~ &~~57.8~~~    &~~92.7~~~           & ~~91.5~~~          & ~~91.4~~~         &~~84.6~~~         \\
				\bottomrule

	\end{tabular}
	\caption{GLUE results~(average of 3 runs) of different compression methods applied on BERT-base. The best results are \textbf{bolded}~($p$-value~<~0.05). The numbers in the parenthesis are training data sizes for each task.}
	\label{table:all}
\end{table*}

\subsubsection{Training Details}
The sparsity-relevant hyperparameter $v$  of step-1 is tuned for each task. We search $p_{init}$ in \{0.7, 0.5, 0.3\} and decay it to zero after half of the total training steps. During training, we
fix the batch size to 32. The max input length is set to 384 for SQuAD and 128 for other tasks. We use the AdamW~\cite{adamw} optimizer and search learning rate in \{2e-5, 3e-5\}. We follow the official implementation of all compared baselines and run structured pruning and matrix factorization methods with a unified logits distillation objective for a fair comparison.

\subsubsection{Compression Setting}

We compress a BERT-base with 86M parameters into compact models of various sizes. 
We refer to BERT-base compressed by  LPAF with preserved rank $k$ as LPAF-$k$. We select $k$ from \{260, 130, 80\}, which corresponds to \{50\%, 25\%, 16\%\} of original parameters. We use  Facebook fvcore to compute FLOPs to measure the computation cost based on the number of floating-point operations for processing one sample. See \tabref{table:stats} for details. We set the number of layers in distillation baselines to \{6, 3, 2\} and tune the sparsity-relevant hyperparameters in structured pruning baselines such that  their final remaining parameters corresponds to \{50\%, 25\%, 16\%\} of BERT-base's parameters and the FLOPs roughly equal to LPAF-\{260, 130, 80\}.


\subsection{Main Results}

\label{sec:main}

\tabref{table:all} and \tabref{table:squad} summarize the overall results on GLUE and SQuAD. Under 50\% parameter budget, as the previous state-of-the-arts algorithms in task-specific distillation and structured pruning, CKD, MetaDistil, and CoFi delivers the strongest performance on certain GLUE tasks~(i.e., RTE, CoLA, SST-2) respectively, while our method performs the best on the others. As the compression rate increases, all distillation methods suffer from more evident accuracy declines compared to structured pruning and matrix factorization methods, which suggests the difficulty of knowledge transfer when the capacity of the student model is small. Compared with ISP and CoFi which remove entire attention heads and neurons, LPAF operates at a finer-grained matrix level and is therefore more flexible. Compared with FLOP and BP$_{\text{hybrid}}$ which remove rank-1 component or consecutive blocks of weight matrices, LPAF can effectively utilize the accurate low-rank subnetwork identified by UP$_{\text{first}}$ and maximally recover task accuracy via the proposed optimizations . Through controlled ablation, we show that low-rank sparsity~(step-1) plays the most critical role in preserving task accuracy, while sparsity-aware SVD and mixed-rank fine-tuning also yield consistent improvements via more accurate sparse matrix approximation and regularized training.

\begin{table}[t]
	\centering
	\scriptsize
	\begin{tabular}{l|cc}
		\toprule
		Task &\textbf{SQuAD v1.1~(88K) }&\textbf{SQuAD v2.0~(131K)  }                                           \\
		\midrule
		\% Params.   &50\% ~25\%  ~16\% &50\% ~25\%  ~16\%    \\
				\hline
		DistilBERT  &85.8~ 78.0~ 66.5 &68.2~ 62.5~ 56.2    \\
		TinyBERT    &82.5~ 58.0~ 38.1 &72.2~ 85.3~ 78.2         \\
				\hline
		Theseus &84.2~ 72.7~ 63.2 &71.2~ 77.2~ 72.8                                    \\
				\hline
		ISP  &86.0~ 84.9~ 81.9 &76.9~ 74.1~ 71.8                                            \\
		FLOP  &88.1~ 85.7~ 81.5 &77.7~ 75.3~ 71.3                                            \\
		CoFi  &87.7~ 86.8~ 84.9 &77.3~ 73.9~ 72.4                                               \\
		
				\hline
		SVD$_{\text{Ft}}$  &87.8~ 85.5~ 81.1 &77.4~ 70.1~ 70.0     \\
		
		LPAF~(ours)  &\textbf{89.1}~ \textbf{87.2}~ \textbf{85.7} &\textbf{79.1}~ \textbf{77.2}~ \textbf{75.1}  \\
		\midrule
		BERT-base  &~~88.2~~~ &~~77.9~~~  \\
		\bottomrule
	\end{tabular}
	\caption{SQuAD results~(average of 3 runs) of different compression methods applied on BERT-base. The best results~($p$-value ~<~0.05) are \textbf{bolded}.}
	\label{table:squad}
\end{table}
\subsection{Analysis}
\label{sec:analysis}

\begin{table}[t]
	\centering
	\scriptsize
	\begin{tabular}{ccccc}
		\toprule
		\multicolumn{2}{c|}{T$_{\text{sparse}}$}    & \multicolumn{3}{c}{LPAF}                                                   \\ 
		\midrule
		$v$  & \multicolumn{1}{c|}{rank} & $k$=260 & $k$=130                        & $k$=80                          \\ 
		\midrule
		0.50 & \multicolumn{1}{c|}{705}  & \textbf{91.3}    & 89.9                           & 86.8                            \\
		0.25 & \multicolumn{1}{c|}{557}  & 91.1    & \textbf{90.1} & 87.2                            \\
		0.10 & \multicolumn{1}{c|}{377}  & 89.7    & 89.5                           & \textbf{89.3 } \\ 
		\bottomrule
	\end{tabular}
   \caption{Effect of different T$_\text{sparse}$  on SST-2.}
   \label{table:diffsparse}
\end{table}
\paragraph{Effect of different $T_\text{sparse}$} 	

 We analyze how different $T_\text{sparse}$ impact the final task performance of LPAF without sparsity-aware SVD and mixed-rank fine-tuning.
The results on SST-2 are summarized in \tabref{table:diffsparse}. As we decrease $v$, $T_\text{sparse}$ becomes more sparse and its rank also monotonically decreases. We observe that for a fixed $k$, the performance of LPAF-$k$ resembles a unimodal distribution of the rank of $T_\text{sparse}$: as the rank gets too high, the increased approximation error overturns the benefit of improved accuracy; when the rank is too low, the drop of accuracy also overturns the benefit of decreased approximation error. Generally, the best performance of LPAF-$k$ for a larger $k$ is achieved at a higher rank of $T_\text{sparse}$ compared to that of a smaller $k$.

\paragraph{Effect of sparsity-aware SVD}

In our sparsity-aware SVD, the reconstruction error of each parameter $\bm{W}_{i,j}$ is weighted by its importance score $\bm{S}_{ij}$. To examine its effectiveness in factorizing sparse matrix, we experiment with two variants on SST-2 dataset: (1) $\bm{S}$ is replaced by coarse-grained binary score $\bm{M}$; (2) non-weighted vanilla SVD.

\begin{table}[t]
	\centering
	\scriptsize
	\begin{tabular}{l|lll}
\toprule
		& \multicolumn{3}{c}{Before Step-3 $\rightarrow$ After Step-3}               \\ 
		\midrule
		Weighting Strategy & \multicolumn{1}{c}{$k$=260} & \multicolumn{1}{c}{$k$=130} & \multicolumn{1}{c}{$k$=130} \\ 
		\midrule
		\multicolumn{1}{c|}{w/ $\bm{S}$}    & \multicolumn{1}{l}{\textbf{81.4} $\rightarrow$ \textbf{92.4}} & \multicolumn{1}{l}{\textbf{79.9} $\rightarrow$ \textbf{90.7}} & \textbf{77.5 }$\rightarrow$ \textbf{89.7} \\ 
		\multicolumn{1}{c|}{w/ $\bm{M}$}    & \multicolumn{1}{l}{81.0 $\rightarrow$ 92.1 }     & \multicolumn{1}{l}{79.7 $\rightarrow$ 90.4}     & 77.2 $\rightarrow$ 89.3     \\ 
		\multicolumn{1}{c|} {vanilla SVD} & \multicolumn{1}{l}{79.1 $\rightarrow$ 91.4} & \multicolumn{1}{l}{77.9 $\rightarrow$ 89.2} & 75.9 $\rightarrow$ 88.8 \\ 
\bottomrule
	\end{tabular}
	\caption{Ablation of sparsity-aware  SVD on SST-2.}
	\label{table:diffsvd}
\end{table}

 In \tabref{table:diffsvd} we show that by informing the sparse matrix factorization process with importance score, 
 more task-relevant information can be retained at the beginning~(Step-2).  After further re-training,  weighting by importance score yields the best results under all choices of $k$, and a simple binary weighting strategy using $\bm{M}$ also brings improvement compared to vanilla SVD. This means that our sparsity-aware SVD is still applicable even when $\bm{S}$ is unavailable.

\paragraph{Effect of mixed-rank fine-tuning} In \tabref{table:wwomixedrank}, we examine the effectiveness of mixed-rank fine-tuning by ablation study.  Results show that mixed-ranking fine-tuning consistently brings improvement over standard fine-tuning under all choices of $k$. Adding the consistency objective $\mathcal{L}_{c}$  stabilizes training and leads to further improvement. 
\begin{table}[t]
	\centering
	\scriptsize
	\begin{tabular}{l|lll}
		\toprule
		Fine-tuning Method & $k$=260& $k$=130 & $k$=80   \\
		\midrule
		mixed-rank &\textbf{92.4}     & \textbf{90.7}   & \textbf{89.7}  \\
		- w/o $\mathcal{L}_{c}$  &91.9  &89.8   &89.1  \\
		\midrule
		vanilla fine-tuning &91.4 & 89.5   &88.8  	 \\
		\bottomrule
	\end{tabular}
	\caption{Ablation of mixed-rank fine-tuning on SST-2.}
	\label{table:wwomixedrank}
\end{table}

We also study the effect of using different values of $p_\text{init}$ on the performance of mixed-rank fine-tuning. From \tabref{table:diffp} we can see that: (1) for $T_\text{factorized}$ with smaller $k$, it prefers a relatively large $p_\text{init}$ because its model capacity is largely reduced and it can benefit more from mixed-ranking fine-tuning to improve generalization; (2) for $T_\text{factorized}$ with larger $k$, a smaller $p_\text{init}$ is more favorable because its higher capacity makes it less likely to converge into bad local minimum; (3) setting $p_\text{init}$ to zero makes our method loses regularization effect brought by gradient-level interaction between factorized sub-matrices and original sparse matrix, thus degenerating performance under all compression ratios.
\begin{table}[t]
	\centering
	\scriptsize
	\begin{tabular}{cl|lll}
		\toprule
		&$p_\text{init}$ & $k$=260     & $k$=130 & $k$=80   \\
		\midrule
		 & 0.7 & 92.1 & 90.2  & \textbf{89.7}  \\
		& 0.5& 92.1 & 90.5   & 89.5  \\
		& 0.3& 92.2 & \textbf{90.7}   & 89.0 \\
		& 0.1& \textbf{92.4} & 90.6   & 89.0 \\
		& 0.0  & 91.8   & 90.0   & 89.3 \\
		\bottomrule
	\end{tabular}
	\caption{Ablation of  different $p_\text{init}$ on SST-2.
	}
	\label{table:diffp}
\end{table}

\subsection{Applicability to Other PLMs}
To verify the general utility of LPAF,  we apply  it to compress an already compact 12-layer and 384-dimensional MiniLM~\cite{minilm} model with 21.5M parameters into 50\% of original parameters and FLOPs. The results are shown in \tabref{table:roberta}. For LPAF, we observe a similar low-rank phenomenon~(281 on average) in the sparse model, demonstrating the general low-rank sparse pattern induced by UP$_\text{first}$. LPAF performs better than or on par with SVD$_{\text{Ft}}$ and the strong distillation method CKD on three representative GLUE tasks, which confirms its general applicability to pre-trained language models of different scales.
%

\begin{table}[t]
	\centering
	\scriptsize
	\begin{tabular}{c|ccc}
		\toprule
		Task & \textbf{SST-2}   & \textbf{QNLI}     & \textbf{MNLI-m/mm}         \\
		\midrule
		CKD       & \textbf{91.2}          & 89.3          & 83.0/83.7        \\
		
		\midrule
		SVD$_{\text{Ft}}$           & 90.0          & 89.6          & 82.8/83.0        \\
		LPAF~(ours)                  &91.1          & \textbf{90.5}          & \textbf{84.4/84.5}          \\
		\midrule
		MiniLM           & 92.4          & 91.2          & 85.0/85.2         \\
		\bottomrule
	\end{tabular}
	\caption{Results~(average of 3 runs) of  compressing MiniLM. Best results are \textbf{bolded}~($p$-value<0.05).}
	\label{table:roberta}
\end{table}

	\section{Conclusion}
We discover that the full-rankness of fine-tuned PLMs is the fundamental bottleneck for the failure of matrix factorization. As a remedy, we employ first-order unstructured pruning to extract the low-rank subnetwork for further factorization. We then propose sparsity-aware SVD and mixed-rank fine-tuning as two optimizations to boost the compression performance. Thorough experiments demonstrate that LPAF can achieve better accuracy-compression trade-offs against existing approaches.

\section*{Limitations}
LPAF bears certain extra training overhead compared to vanilla fine-tuning. Specifically, during first-order pruning procedure, at certain iterations we need to rank all model parameters according to their importance 
scores. Taking the soring time into account, the pruning process takes about 1.15x times compared to fine-tuning. In the last stage of LPAF, we perform mixed-rank fine-tuning as an effective regularization scheme to 
facilitate the generalization ability of the model being compressed. Because tach mini-batch data samples will be fed to the model twice, LPAF takes approximatedly 1.4x memory and 1.3x time v.s. vanilla fine-tuning. Nonetheless, we believe it is worthwhile since we only need to do it once and the compressed model can be deployed anywhere needed.

	\bibliography{acl2023}

\begin{thebibliography}{33}
\expandafter\ifx\csname natexlab\endcsname\relax\def\natexlab#1{#1}\fi

\bibitem[{Ben~Noach and Goldberg(2020)}]{svd}
Matan Ben~Noach and Yoav Goldberg. 2020.
\newblock \href {https://aclanthology.org/2020.aacl-main.88} {Compressing
  pre-trained language models by matrix decomposition}.
\newblock In \emph{Proceedings of the 1st Conference of the Asia-Pacific
  Chapter of the Association for Computational Linguistics and the 10th
  International Joint Conference on Natural Language Processing}, pages
  884--889, Suzhou, China. Association for Computational Linguistics.

\bibitem[{Bengio et~al.(2013)Bengio, L{\'{e}}onard, and Courville}]{st}
Yoshua Bengio, Nicholas L{\'{e}}onard, and Aaron~C. Courville. 2013.
\newblock \href {http://arxiv.org/abs/1308.3432} {Estimating or propagating
  gradients through stochastic neurons for conditional computation}.
\newblock \emph{CoRR}, abs/1308.3432.

\bibitem[{Chen et~al.(2018)Chen, Si, Li, Chelba, and Hsieh}]{group}
Patrick~H. Chen, Si~Si, Yang Li, Ciprian Chelba, and Cho{-}Jui Hsieh. 2018.
\newblock \href {http://arxiv.org/abs/1806.06950} {Groupreduce: Block-wise
  low-rank approximation for neural language model shrinking}.
\newblock \emph{CoRR}, abs/1806.06950.

\bibitem[{Chen et~al.(2020)Chen, Frankle, Chang, Liu, Zhang, Wang, and
  Carbin}]{chen2020lottery}
Tianlong Chen, Jonathan Frankle, Shiyu Chang, Sijia Liu, Yang Zhang, Zhangyang
  Wang, and Michael Carbin. 2020.
\newblock \href {http://arxiv.org/abs/2007.12223} {The lottery ticket
  hypothesis for pre-trained bert networks}.

\bibitem[{Devlin et~al.(2019)Devlin, Chang, Lee, and Toutanova}]{bert}
Jacob Devlin, Ming-Wei Chang, Kenton Lee, and Kristina Toutanova. 2019.
\newblock \href {https://doi.org/10.18653/v1/N19-1423} {{BERT}: Pre-training of
  deep bidirectional transformers for language understanding}.
\newblock In \emph{Proceedings of the 2019 Conference of the North {A}merican
  Chapter of the Association for Computational Linguistics: Human Language
  Technologies, Volume 1 (Long and Short Papers)}, pages 4171--4186,
  Minneapolis, Minnesota. Association for Computational Linguistics.

\bibitem[{Dolan and Brockett(2005)}]{mrpc}
William~B. Dolan and Chris Brockett. 2005.
\newblock \href {https://aclanthology.org/I05-5002} {Automatically constructing
  a corpus of sentential paraphrases}.
\newblock In \emph{Proceedings of the Third International Workshop on
  Paraphrasing ({IWP}2005)}.

\bibitem[{Han et~al.(2015)Han, Pool, Tran, and Dally}]{mag}
Song Han, Jeff Pool, John Tran, and William~J. Dally. 2015.
\newblock \href {http://arxiv.org/abs/1506.02626} {Learning both weights and
  connections for efficient neural networks}.
\newblock \emph{CoRR}, abs/1506.02626.

\bibitem[{Hsu et~al.(2021)Hsu, Hua, Chang, Lou, Shen, and
  Jin}]{hsu2021language}
Yen-Chang Hsu, Ting Hua, Sungen Chang, Qian Lou, Yilin Shen, and Hongxia Jin.
  2021.
\newblock Language model compression with weighted low-rank factorization.
\newblock In \emph{International Conference on Learning Representations}.

\bibitem[{Jiao et~al.(2020)Jiao, Yin, Shang, Jiang, Chen, Li, Wang, and
  Liu}]{tinybert}
Xiaoqi Jiao, Yichun Yin, Lifeng Shang, Xin Jiang, Xiao Chen, Linlin Li, Fang
  Wang, and Qun Liu. 2020.
\newblock \href {https://doi.org/10.18653/v1/2020.findings-emnlp.372}
  {{T}iny{BERT}: Distilling {BERT} for natural language understanding}.
\newblock In \emph{Findings of the Association for Computational Linguistics:
  EMNLP 2020}, pages 4163--4174, Online. Association for Computational
  Linguistics.

\bibitem[{Lagunas et~al.(2021)Lagunas, Charlaix, Sanh, and Rush}]{block}
Fran{\c{c}}ois Lagunas, Ella Charlaix, Victor Sanh, and Alexander Rush. 2021.
\newblock \href {https://doi.org/10.18653/v1/2021.emnlp-main.829} {Block
  pruning for faster transformers}.
\newblock In \emph{Proceedings of the 2021 Conference on Empirical Methods in
  Natural Language Processing}, pages 10619--10629, Online and Punta Cana,
  Dominican Republic. Association for Computational Linguistics.

\bibitem[{Liu et~al.(2019)Liu, Ott, Goyal, Du, Joshi, Chen, Levy, Lewis,
  Zettlemoyer, and Stoyanov}]{roberta}
Yinhan Liu, Myle Ott, Naman Goyal, Jingfei Du, Mandar Joshi, Danqi Chen, Omer
  Levy, Mike Lewis, Luke Zettlemoyer, and Veselin Stoyanov. 2019.
\newblock \href {http://arxiv.org/abs/1907.11692} {Roberta: {A} robustly
  optimized {BERT} pretraining approach}.
\newblock \emph{CoRR}, abs/1907.11692.

\bibitem[{Loshchilov and Hutter(2017)}]{adamw}
Ilya Loshchilov and Frank Hutter. 2017.
\newblock \href {http://arxiv.org/abs/1711.05101} {Fixing weight decay
  regularization in adam}.
\newblock \emph{CoRR}, abs/1711.05101.

\bibitem[{Louizos et~al.(2018)Louizos, Welling, and Kingma}]{l0}
Christos Louizos, Max Welling, and Diederik~P Kingma. 2018.
\newblock Learning sparse neural networks through $ l\_0 $ regularization.
\newblock \emph{arXiv preprint arXiv:1712.01312}.

\bibitem[{Molchanov et~al.(2016)Molchanov, Tyree, Karras, Aila, and
  Kautz}]{isp}
Pavlo Molchanov, Stephen Tyree, Tero Karras, Timo Aila, and Jan Kautz. 2016.
\newblock \href {http://arxiv.org/abs/1611.06440} {Pruning convolutional neural
  networks for resource efficient transfer learning}.
\newblock \emph{CoRR}, abs/1611.06440.

\bibitem[{Nakkiran et~al.(2019)Nakkiran, Kaplun, Bansal, Yang, Barak, and
  Sutskever}]{overpara}
Preetum Nakkiran, Gal Kaplun, Yamini Bansal, Tristan Yang, Boaz Barak, and Ilya
  Sutskever. 2019.
\newblock \href {http://arxiv.org/abs/1912.02292} {Deep double descent: Where
  bigger models and more data hurt}.
\newblock \emph{CoRR}, abs/1912.02292.

\bibitem[{Park et~al.(2021)Park, Kim, and Yang}]{CKD}
Geondo Park, Gyeongman Kim, and Eunho Yang. 2021.
\newblock \href {http://arxiv.org/abs/2109.08359} {Distilling linguistic
  context for language model compression}.
\newblock \emph{CoRR}, abs/2109.08359.

\bibitem[{Rajpurkar et~al.(2018)Rajpurkar, Jia, and Liang}]{squadv2.0}
Pranav Rajpurkar, Robin Jia, and Percy Liang. 2018.
\newblock \href {http://arxiv.org/abs/1806.03822} {Know what you don't know:
  Unanswerable questions for squad}.
\newblock \emph{CoRR}, abs/1806.03822.

\bibitem[{Rajpurkar et~al.(2016)Rajpurkar, Zhang, Lopyrev, and
  Liang}]{qnliandsquad}
Pranav Rajpurkar, Jian Zhang, Konstantin Lopyrev, and Percy Liang. 2016.
\newblock \href {http://arxiv.org/abs/1606.05250} {Squad: 100, 000+ questions
  for machine comprehension of text}.
\newblock \emph{CoRR}, abs/1606.05250.

\bibitem[{Sanh et~al.(2019)Sanh, Debut, Chaumond, and Wolf}]{distilbert}
Victor Sanh, Lysandre Debut, Julien Chaumond, and Thomas Wolf. 2019.
\newblock \href {http://arxiv.org/abs/1910.01108} {Distilbert, a distilled
  version of {BERT:} smaller, faster, cheaper and lighter}.
\newblock \emph{CoRR}, abs/1910.01108.

\bibitem[{Sanh et~al.(2020)Sanh, Wolf, and Rush}]{movement}
Victor Sanh, Thomas Wolf, and Alexander~M. Rush. 2020.
\newblock \href {http://arxiv.org/abs/2005.07683} {Movement pruning: Adaptive
  sparsity by fine-tuning}.
\newblock \emph{CoRR}, abs/2005.07683.

\bibitem[{Socher et~al.(2013)Socher, Perelygin, Wu, Chuang, Manning, Ng, and
  Potts}]{sst2}
Richard Socher, Alex Perelygin, Jean Wu, Jason Chuang, Christopher~D. Manning,
  Andrew Ng, and Christopher Potts. 2013.
\newblock \href {https://aclanthology.org/D13-1170} {Recursive deep models for
  semantic compositionality over a sentiment treebank}.
\newblock In \emph{Proceedings of the 2013 Conference on Empirical Methods in
  Natural Language Processing}, pages 1631--1642, Seattle, Washington, USA.
  Association for Computational Linguistics.

\bibitem[{Srebro and Jaakkola(2003)}]{weightedsvd}
Nathan Srebro and Tommi Jaakkola. 2003.
\newblock Weighted low-rank approximations.
\newblock In \emph{Proceedings of the 20th international conference on machine
  learning (ICML-03)}, pages 720--727.

\bibitem[{Stewart(1998)}]{bestsvd}
Gilbert~W Stewart. 1998.
\newblock Perturbation theory for the singular value decomposition.
\newblock Technical report.

\bibitem[{Sun et~al.(2019)Sun, Cheng, Gan, and Liu}]{pkd}
Siqi Sun, Yu~Cheng, Zhe Gan, and Jingjing Liu. 2019.
\newblock \href {http://arxiv.org/abs/1908.09355} {Patient knowledge
  distillation for {BERT} model compression}.
\newblock \emph{CoRR}, abs/1908.09355.

\bibitem[{Vaswani et~al.(2017)Vaswani, Shazeer, Parmar, Uszkoreit, Jones,
  Gomez, Kaiser, and Polosukhin}]{transformer}
Ashish Vaswani, Noam Shazeer, Niki Parmar, Jakob Uszkoreit, Llion Jones,
  Aidan~N. Gomez, Lukasz Kaiser, and Illia Polosukhin. 2017.
\newblock \href {http://arxiv.org/abs/1706.03762} {Attention is all you need}.
\newblock \emph{CoRR}, abs/1706.03762.

\bibitem[{Wang et~al.(2018)Wang, Singh, Michael, Hill, Levy, and Bowman}]{glue}
Alex Wang, Amanpreet Singh, Julian Michael, Felix Hill, Omer Levy, and
  Samuel~R. Bowman. 2018.
\newblock \href {http://arxiv.org/abs/1804.07461} {{GLUE:} {A} multi-task
  benchmark and analysis platform for natural language understanding}.
\newblock \emph{CoRR}, abs/1804.07461.

\bibitem[{Wang et~al.(2020{\natexlab{a}})Wang, Wei, Dong, Bao, Yang, and
  Zhou}]{minilm}
Wenhui Wang, Furu Wei, Li~Dong, Hangbo Bao, Nan Yang, and Ming Zhou.
  2020{\natexlab{a}}.
\newblock Minilm: Deep self-attention distillation for task-agnostic
  compression of pre-trained transformers.
\newblock \emph{Advances in Neural Information Processing Systems},
  33:5776--5788.

\bibitem[{Wang et~al.(2020{\natexlab{b}})Wang, Wohlwend, and Lei}]{flop}
Ziheng Wang, Jeremy Wohlwend, and Tao Lei. 2020{\natexlab{b}}.
\newblock \href {https://doi.org/10.18653/v1/2020.emnlp-main.496} {Structured
  pruning of large language models}.
\newblock In \emph{Proceedings of the 2020 Conference on Empirical Methods in
  Natural Language Processing (EMNLP)}, pages 6151--6162, Online. Association
  for Computational Linguistics.

\bibitem[{Williams et~al.(2017)Williams, Nangia, and Bowman}]{mnli}
Adina Williams, Nikita Nangia, and Samuel~R. Bowman. 2017.
\newblock \href {http://arxiv.org/abs/1704.05426} {A broad-coverage challenge
  corpus for sentence understanding through inference}.
\newblock \emph{CoRR}, abs/1704.05426.

\bibitem[{Xia et~al.(2022)Xia, Zhong, and Chen}]{sp-l0}
Mengzhou Xia, Zexuan Zhong, and Danqi Chen. 2022.
\newblock \href {https://doi.org/10.18653/v1/2022.acl-long.107} {Structured
  pruning learns compact and accurate models}.
\newblock In \emph{Proceedings of the 60th Annual Meeting of the Association
  for Computational Linguistics (Volume 1: Long Papers)}, pages 1513--1528,
  Dublin, Ireland. Association for Computational Linguistics.

\bibitem[{Xu et~al.(2020)Xu, Zhou, Ge, Wei, and Zhou}]{theseus}
Canwen Xu, Wangchunshu Zhou, Tao Ge, Furu Wei, and Ming Zhou. 2020.
\newblock \href {http://arxiv.org/abs/2002.02925} {Bert-of-theseus: Compressing
  {BERT} by progressive module replacing}.
\newblock \emph{CoRR}, abs/2002.02925.

\bibitem[{Xu et~al.(2021)Xu, Luo, Wang, Chang, Huang, Huang, and Huang}]{cap}
Runxin Xu, Fuli Luo, Chengyu Wang, Baobao Chang, Jun Huang, Songfang Huang, and
  Fei Huang. 2021.
\newblock \href {http://arxiv.org/abs/2112.07198} {From dense to sparse:
  Contrastive pruning for better pre-trained language model compression}.
\newblock \emph{CoRR}, abs/2112.07198.

\bibitem[{Zhou et~al.(2022)Zhou, Xu, and McAuley}]{metadistil}
Wangchunshu Zhou, Canwen Xu, and Julian McAuley. 2022.
\newblock \href {https://aclanthology.org/2022.acl-long.485} {{BERT} learns to
  teach: Knowledge distillation with meta learning}.
\newblock In \emph{Proceedings of the 60th Annual Meeting of the Association
  for Computational Linguistics (Volume 1: Long Papers)}, pages 7037--7049,
  Dublin, Ireland. Association for Computational Linguistics.

\end{thebibliography}
	\bibliographystyle{acl_natbib}
	\newpage
	\appendix
	\section{Preliminary Study}
\label{sec:A}
We show the accuracy-rank trade-offs on MRPC, RTE, and CoLA in \figref{fig:pre}~(CoLA is additionally included compared to the main body of the paper). The observation on CoLA is similar to MRPC/RTE: first-order unstructured pruning can extract subnetworks that are most accurate while having the lowest average matrix rank, which lays the crucial foundation of later factorization.

\begin{figure*}[t]
	\centering
	\scalebox{0.285}{\includegraphics{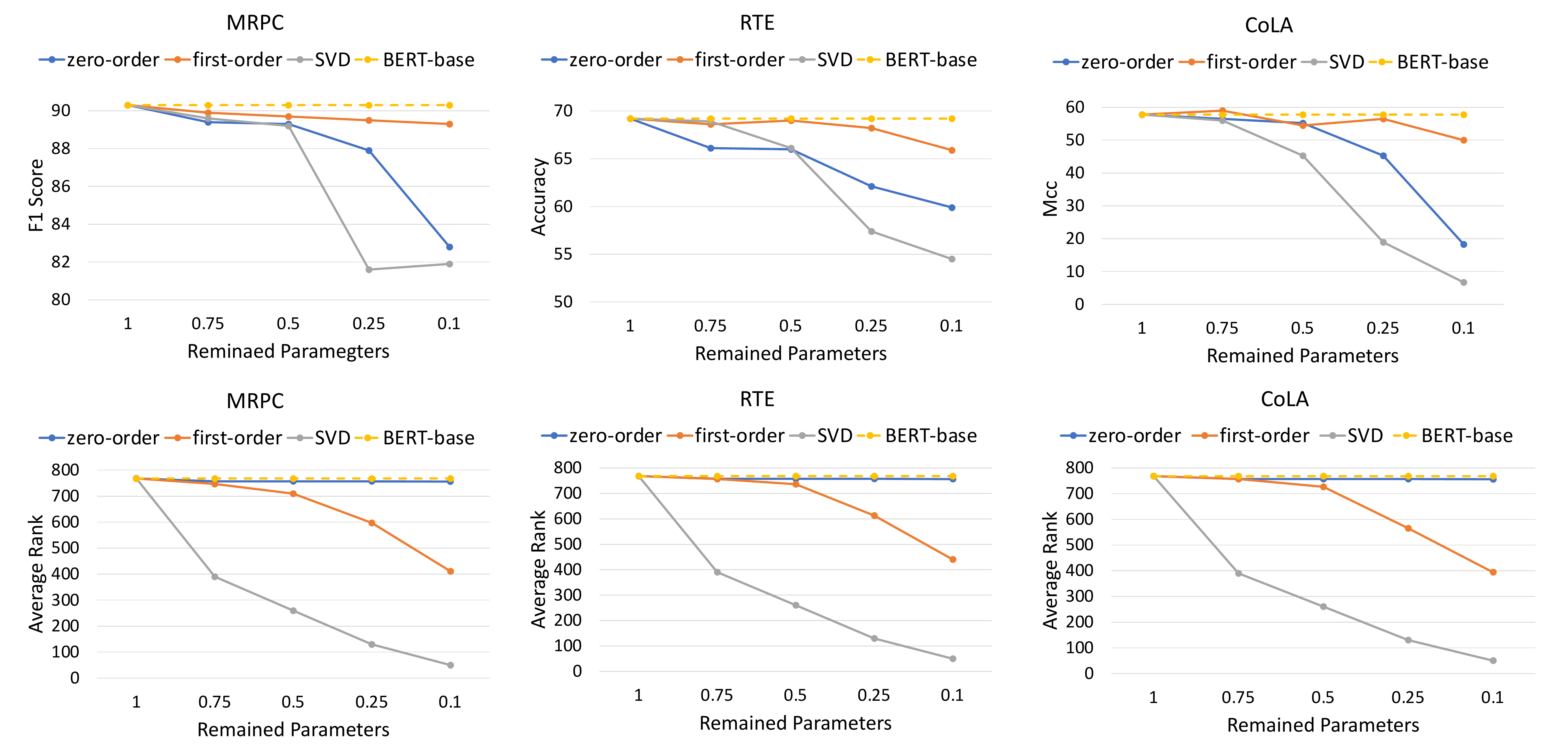}}
	\caption{Task accuracy (top half) and average matrix rank (bottom half) v.s. percentage of original parameters retained. The dashed line indicates the performance/rank upper bound by fine-tuning the full-scale BERT-base model.}
	\label{fig:pre}
\end{figure*}
	\clearpage
\end{document}